\documentclass[conference]{IEEEtran}
\IEEEoverridecommandlockouts
\usepackage{cite}
\usepackage{amsmath,amssymb,amsfonts}
\usepackage{algorithmic}
\usepackage{graphicx}
\usepackage{textcomp}
\usepackage{longtable}
\usepackage{xcolor}
\usepackage{amsmath}
\usepackage{blindtext, subfig}
\usepackage{stfloats} 
\usepackage{booktabs} 
\usepackage[utf8]{inputenc}
\def\BibTeX{{\rm B\kern-.05em{\sc i\kern-.025em b}\kern-.08em
    T\kern-.1667em\lower.7ex\hbox{E}\kern-.125emX}}
\begin{document}

\title{Quantification of Uncertainties in Deep Learning - based Environment Perception\\}

\author{\IEEEauthorblockN{1\textsuperscript{st} Marco Braun}
\IEEEauthorblockA{\textit{University of Wuppertal} \\
Wuppertal, Germany \\
marco.braun@aptiv.com}
\and
\IEEEauthorblockN{2\textsuperscript{nd} Moritz Luszek}
\IEEEauthorblockA{\textit{Aptiv} \\
Wuppertal, Germany \\
moritz.luszek@aptiv.com}
\and
\IEEEauthorblockN{3\textsuperscript{rd} Jan Siegemund}
\IEEEauthorblockA{\textit{Aptiv} \\
Wuppertal, Germany \\
jan.siegemund@aptiv.com}
\and
\IEEEauthorblockN{4\textsuperscript{th} Kevin Kollek}
\IEEEauthorblockA{\textit{University of Wuppertal} \\
Wuppertal, Germany \\
kollek@uni-wuppertal.de}
\and
\IEEEauthorblockN{5\textsuperscript{th} Anton Kummert}
\IEEEauthorblockA{\textit{University of Wuppertal} \\
Wuppertal, Germany \\
kummert@uni-wuppertal.de}
}


\makeatletter
\def\ps@IEEEtitlepagestyle{
  \def\@oddfoot{\mycopyrightnotice}
  \def\@evenfoot{}
}
\def\mycopyrightnotice{
  {\footnotesize
  \begin{minipage}{\textwidth}
  \centering
  Copyright~\copyright~2021 IEEE.  Personal use of this material is permitted.  Permission from IEEE must be obtained for all other uses, in any current or future media, including reprinting/republishing this material for advertising or promotional purposes, creating new collective works, for resale or redistribution to servers or lists, or reuse of any copyrighted component of this work in other works. DOI: 10.1109/COINS51742.2021.9524106
  \end{minipage}
  }
}

\maketitle

\IEEEpubidadjcol

\begin{abstract}
In this work, we introduce a novel Deep Learning-based method to perceive the environment of a vehicle based on radar scans while accounting for uncertainties in its predictions. The environment of the host vehicle is segmented into equally sized grid cells which are classified individually. Complementary to the segmentation output, our Deep Learning-based algorithm is capable of differentiating uncertainties in its predictions as being related to an inadequate model (epistemic uncertainty) or noisy data (aleatoric uncertainty). To this end, weights are described as probability distributions accounting for uncertainties in the model parameters. Distributions are learned in a supervised fashion using gradient descent. We prove that uncertainties in the model output correlate with the precision of its predictions. Compared to previous concepts, we show superior performance of our approach to reliably perceive the environment of a vehicle. \\
\end{abstract}

\begin{IEEEkeywords}
Deep Learning, Environment Perception, Uncertainty Estimation \\
\end{IEEEkeywords}

\section{Introduction}

In recent years, the development of advanced driver assistance systems up to fully autonomous driving has been driven forward extensively. To implement these systems, sensors such as cameras, radar or lidar are utilized to perceive the surroundings of a vehicle. While camera and lidar sensors tend to fail in challenging weather conditions like fog or heavy rain, radar sensors show superior reliability in these situations while being less expensive. Moreover,  reflections from radar sensors contain valuable intrinsic properties such as the relative velocity of the detected object, also known as Doppler velocity $v_{R}$, and the reflection intensity ($RCS$). In a partially static and dynamic environment, radar returns are therefore particularly suitable to derive scene information. This perception can then be used in a variety of driving assistance functions such as collision avoidance or ego trajectory calculation in automated driving.\\
Inverse Sensor Models (ISM) \cite{probabilistic_robotics} have been extensively used to segment the surroundings of a vehicle into drivable and occupied areas based on radar signals. By defining occupancy grid maps (OGM) as a semantic segmentation task, latest approaches such as \cite{occupancy_grid_semseg} outperformed traditional ISM based methods. These algorithms are trained by utilizing deep learning (DL) techniques to predict whether a cell is occupied or free based on radar returns.\\
\begin{figure}
\centering
\includegraphics[width=58mm, trim={0 0 14.5cm 0.0cm}, clip]{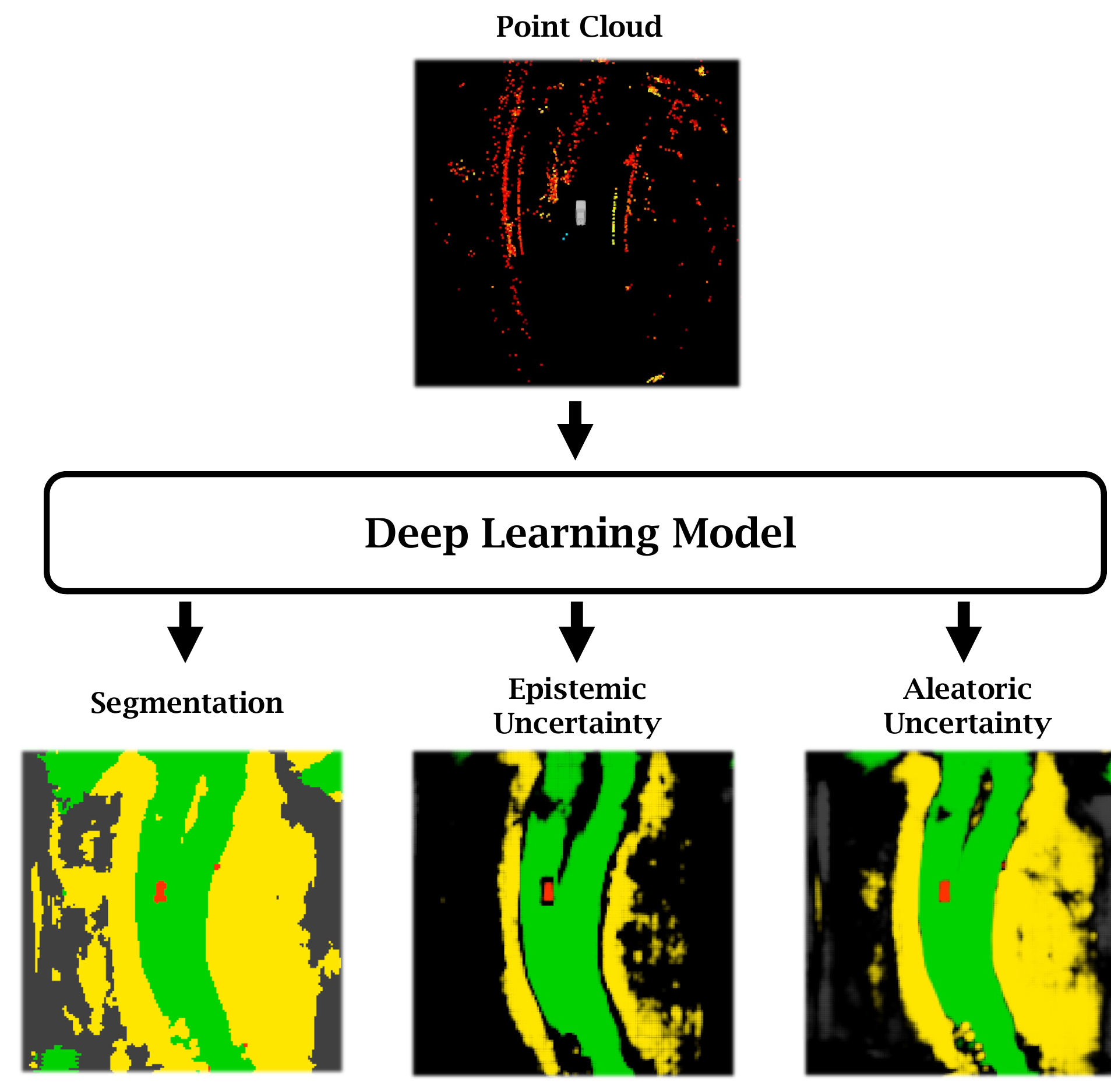}
\caption{\textbf{System Design}: Radar reflections are processed by a Deep Learning system to predict the environment around the vehicle. This environment is partitioned into 160 x 160 grid cells of size 0.5$m^{2}$ with the vehicle located in the center. The system classifies each cell individually as \textit{free} (green), \textit{occupied} (yellow), \textit{moving object} (red) or \textit{unknown} (gray). In addition, the system quantifies epistemic and aleatoric uncertainties associated with the prediction of each cell. Uncertainties in the predictions are presented by darkened output cells.}
\label{fig:system_overwiew}
\end{figure}
While calculations by ISMs, however, are comprehensible due to their explicitly formulated model and therefore trustworthy, the internal processing of data in a neural network can hardly be interpreted. In most cases, the user has no choice but to blindly trust the network's predictions, with no knowledge about internal reasoning and decision-making. The safety-critical nature of driver assistance systems, however, requires a reliable prediction for perceiving the area around the vehicle.\newline
To address these safety concerns, the system must be able to quantify the uncertainties associated with a prediction from a neural network as depicted in Figure \ref{fig:system_overwiew}. Therefore, measuring two types of uncertainty is desired: \textit{Aleatoric uncertainty} accounts for noise and ambiguities in the data itself. When processing radar reflections, this uncertainty can be caused by spatial inaccuracies of a reflection or a high variance in radar-specific properties that are characteristic for certain classes. \textit{Epistemic uncertainty} represents the uncertainty in the model parameters. It captures how close the characteristics of a given situation match the data on which a network was trained on. By quantifying this uncertainty, the system is able to identify unfamiliar environmental conditions that would otherwise lead to an overconfident prediction.\\
While aleatoric uncertainty can be captured by mapping the network outputs to a categorical distribution \cite{aleatoric_uncertainty_capturing}, estimating the uncertainty in the model parameters themselves requires modifications to the network structure. Epistemic uncertainty in a model can be captured by placing distributions on its weights that theoretically represent all plausible model parameters, given the data \cite{Weight_Uncertainty_in_Neural_Networks}\cite{dropout_as_bayesian_approximation}. Approaches like \cite{Bayesian_SegNet} \cite{Uncertainties_in_CV} present Bayesian Neural Network (BNN) architectures that quantify epistemic uncertainties for the semantic segmentation of images by utilizing Monte Carlo Dropout \cite{dropout_as_bayesian_approximation}. We build up on these methods to reliably perceive the environment of a vehicle based on radar scans.\\
\\
In the first part of this work, we show how Gaussian distributions on the weights can be used to capture model uncertainty and how aleatoric, data-related, uncertainty is derived from the predictions of the same network. Building up on that, we present a model to segment the environment of a vehicle by processing radar data while accounting for uncertainties in its predictions. By parameterizing each weight in this model by a Gaussian distribution, the network doubles in size in terms of trainable parameters. To counteract this, we present a more efficient hybrid deterministic, probabilistic network architecture. In the last section, we analyze quantitatively and qualitatively the ability of our approach to perceive the environment while taking into account epistemic and aleatory uncertainties. Finally, we compare our approach to state-of-the-art implementations based on MC Dropout which we adapt for the task of environment perception.\\

\section{Related Work}\label{sec:related_work}

\subsection{Uncertainties in Semantic Segmentation}
In DL-systems, we want our neural network to learn how to map an input $x \in X$ on a target $y \in Y$. For semantic segmentation of grid cells, the input data shape $X$ corresponds to a regular grid structure of $c_{l}$ \textit{x} $c_{w}$ cells, each containing $F_{in}$ input features. We then apply a network $f_{\theta}(x)$ with weights w that are parameterized by $\theta$ to predict $C$ class probabilities for each cell individually. In order to optimize an ordinary neural network model to perform semantic segmentation, the maximum-likelihood estimate (MLE) is captured by maximizing the log-likelihood 
\begin{equation}\label{eq:mle}
w_{\text{MLE}} = \text{argmax}_{w} \log\:p(Y |X, w)
\end{equation}
of the class probabilities for each cell given the training data $X_{train}$, $Y_{train}$. These networks, however, are not feasible to account for uncertainties in their parameters.\\
\\
To capture epistemic uncertainty, BNN \cite{BNN_0}\cite{BNN_1} architectures can be deployed. These networks are implemented by placing prior probability distributions $p(w)$ on the model weights so that the maximum a posteriori (MAP) objective 
\begin{equation}\label{eq:map}
w_{\text{MAP}} = \text{argmax}_{w} \log\:p(Y |X, w) + \log p(w)
\end{equation}
can be applied to find the optimal distributions on the weights w given the training data $X_{train}$, $Y_{train}$ \cite{Weight_Uncertainty_in_Neural_Networks}. As a result, the posterior probability distribution $p(w|X, Y)$ indicates an infinite set of all possible model parameters of the network given the training data.\\
In real world scenarios, the exact posterior distribution, however, is not tractable and not parameterizable. Therefore, different approaches were developed to approximate $p(w | X, Y)$.
\\
By placing Bernoulli distributions on the weights of a neural network, \cite{dropout_as_bayesian_approximation} presents a BNN that is used in a broad variety of approaches to capture epistemic uncertainty in the semantic segmentation of images. For \textit{Bayesian SegNets} \cite{Bayesian_SegNet}, the authors slightly modify the general idea of BNNs by placing dropout layers between layers of their network that are activated during testing. This approach of capturing the approximate posterior distribution is called \textit{Monte Carlo Dropout (MC Dropout)}. Similarly, the authors of \cite{Uncertainties_in_CV} present a modified \textit{DenseNet} \cite{Densenet} for semantic segmentation of images while capturing epistemic and aleatoric uncertainties. Comparable to \textit{Bayesian SegNet}, epistemic uncertainty is obtained by utilizing dropout layers. Furthermore, the authors of \cite{Uncertainties_in_CV} capture heteroscedastic aleatoric uncertainty - aleatoric uncertainty dependent on the network inputs - by implementing a loss attenuation that is learned in an unsupervised manner. The resulting attenuation factor is then interpreted as the aleatoric uncertainty predicted by the network. \\ 

\subsection{Deep Learning on Environment Perception}

Returns from radar sensors are broadly used for the calculation of occupancy grid maps due to their long range sensing and superior robustness in challenging weather conditions. Traditionally, grid maps are deducted by using a combination of ISMs and Bayesian filtering \cite{probabilistic_robotics} to calculate the probability for a grid cell being \textit{free} or \textit{occupied} based on sensor returns \cite{Ogm_0}\cite{Ogm_1}. Lately, machine learning approaches were developed to perform occupancy grid mapping as a data-driven task. In \cite{occupancy_grid_semseg}, the authors formulate ISMs as a three class semantic segmentation task by applying a neural network to predict for each grid cell whether it is \textit{occupied}, \textit{free} or \textit{unobserved}. Ground truth (GT) is generated from lidar ray tracing so that the system can be trained in a self-supervised manner. This approach outperforms traditional grid mapping techniques.\\
\\
The authors of \cite{probably_unknown} expand the idea of applying a DL-based systems to learn ISMs by modeling heteroscedastic aleatoric uncertainty that, based on the input radar scan, should  indicate whether the system assumes a cell to be occluded. This is achieved by treating each output as a normally distributed latent variable \textit{z}. This variable is parameterized by a standard deviation factor $\gamma_{\phi}$ which accounts for the uncertainty of a cell being observable and a mean value $\mu_{\phi}$ indicating the predicted probability of a cell being occupied.\\ 
\\
In order to produce reliable grid maps based on a neural network that processes sensor scans, capturing the uncertainty that arises from occlusion is not sufficient. Instead, the network's occasional ignorance about how to reason on the environment due to insufficient training data or noise related to sensor-specific properties needs to be quantified. Additionally, the two presented approaches \cite{occupancy_grid_semseg} and \cite{probably_unknown} show that lidar sensors are broadly utilized to gather data about the environment of a vehicle due to their high measurement accuracy. The transition from knowledge gained from the lidar sensor to a system that consumes radar data, however, poses additional challenges related to different sensor characteristics which further reduces the confidence in the network predictions. In the following sections, we therefore define a model for environment perception that is able to reason about the reliability of its outputs by capturing epistemic and aleatoric uncertainties.\\

\section{Method}\label{sec:method}

\subsection{Measuring Uncertainties}\label{sec:measuring_uncertainties}
As stated above, we want to capture both epistemic and aleatoric uncertainties from our network outputs. Differentiating between these two kinds is desired as it opens up opportunities to better interpret each network prediction so that strategies can be deduced on how to handle each situation. In order to measure both epistemic and aleatoric uncertainty, we first calculate the predictive uncertainty ${H}_{p}$ which can then be decomposed. In classification problems this predictive uncertainty can be captured by calculating the entropy of a network output \cite{Deep_Bayesian_Active_learning}. We calculate this predictive entropy as
\begin{equation}\label{eq:predictive_entropy_I}
\mathbb{H}_{p}= - \sum^{C} p(y^{*}|x^{*}, X, Y)\:\log\:p(y^{*}|x^{*}, X, Y)
\end{equation}
for \textit{C} classes. This term can be interpreted as a composition of aleatoric and epistemic uncertainties\cite{depeweg_uncertainty_decomposition}. \\
We can deduce the former by calculating the entropy while assuming a fixed value for the weights of the network
\begin{equation}
\label{eq:aleatoric_entropy}
\mathbb{H}_{a}= E_{q(w | \theta)} [\mathbb{H}(y^{*}|x^{*}, w)]\\
\end{equation}
since the resulting uncertainty $\mathbb{H}_{a}$ arises from the input data rather than the weights \cite{depeweg_uncertainty_decomposition}.\\
The epistemic uncertainty can then be obtained as the difference between the predictive and the aleatoric uncertainty
\begin{equation}\label{eq:epistemic_entropy}
\mathbb{H}_{e} = \mathbb{H}_{p} - \mathbb{H}_{a}
\end{equation}
since it results from uncertainties in the weight parameters that are not covered by $\mathbb{H}_{a}$.

\subsection{Epistemic Uncertainty in environment perception}\label{sec:epistemic_uncertainty}

\begin{figure}[t]
\centering
\includegraphics[width=60mm]{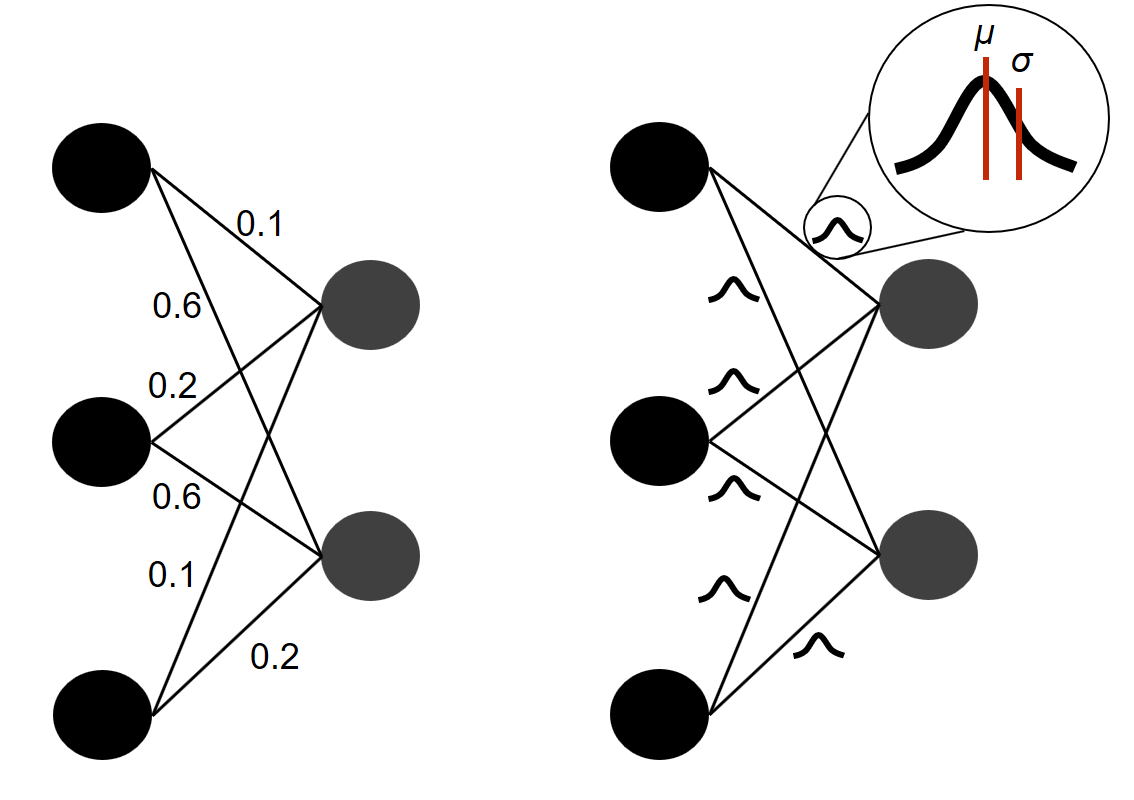}
\caption{While weights in conventional neural networks are defined by point estimates (left side), we replace them by Gaussian distributions $q^{*}(w | \theta)$ that are parameterized by learnable parameters mean $\mu$ and standard deviation $\sigma$ (right side). The distributions on the weights are used to approximate the posterior probability distribution $p(w | X, Y)$}
\label{fig:bnn}
\end{figure}

As mentioned in Section \ref{sec:related_work}, epistemic uncertainty is modeled by capturing the posterior distribution $p(w | X, Y)$ on the weights given the data. This posterior distribution, however, is intractable as it indicates the probability of each weight in a neural network to take any possible value. We therefore approximate $p(w | X, Y)$ by a parameterizable Gaussian distributions $q(w | \theta)$ with parameters $\theta = [\mu, \sigma]$ (Figure \ref{fig:bnn}). These parameters can then be learned by backpropagation using variational inference
\cite{Variational_Approximation_0}\cite{Variational_Inference}\cite{Weight_Uncertainty_in_Neural_Networks}.\\ 
During training, we therefore want to minimize the Kullback-Leibler (KL) divergence between the actual posterior $p(w | X, Y)$ and its parameterized approximation $q(w | \theta)$:
\begin{equation}\label{eq:kl}
\theta^{*} = \text{arg min}_{\theta} \text{KL}[q(w | \theta)||p(w | X, Y)]
\end{equation}
While this objective function still contains the intractable posterior distribution $p(w | X, Y)$, it can be transformed into the Evidence Lower Bound function\cite{Elbo}\cite{Weight_Uncertainty_in_Neural_Networks} which is used to define the loss function of our training
\begin{equation}\label{eq:kl_loss}
L(\theta, X, Y) = \text{KL}[q(w | \theta)||p(w)] - \text{E}_{q(w|\theta)}[\log\:p(Y |X, w)]
\end{equation}
The latter term of equation \ref{eq:kl_loss} represents the data dependent log-likelihood loss. This term indicates how well the network is able to map the input data on the GT. The former part of equation \ref{eq:kl_loss} depends on the prior probability distribution $p(w) = \mathcal{N}(w|0, \gamma\textit{I})$. As default, we set $\gamma$=1. This prior dependent loss increases the variance of distributions on weights that are rarely optimized during training, indicating a high uncertainty in those weight parameters. We then optimize our network by backpropagation utilizing the reparameterization trick \cite{reparameterization_trick_kingma}.
\\
For semantic segmentation, the categorical distribution $p(y^{*}|x^{*}, X, Y)$ for each output cell can then in theory be obtained by marginalizing over the learned distributions on the weights
\begin{equation}\label{eq:marginalization_integral}
p(y^{*}|x^{*}, X, Y) = \int p(y^{*}|x^{*}, w) q(w | \theta)\:dw
\end{equation}
for $y^{*}$ defining the output of the model $f_{\theta}(x^{*})$ based on input data $x^{*}$.\\
In practice, we apply Monte Carlo sampling to approximate this integral from equation \ref{eq:marginalization_integral} by drawing \textit{N} point estimates $w_{n} {\huge\raisebox{-0.9ex}{\~{ }}}$ $q(w | \theta)$ for each weight from the approximated posterior distributions:
\begin{equation}\label{eq:marginalization_approximated}
p(y^{*}|x^{*}, X, Y) \approx \dfrac{1}{N} \sum_{n=1}^{N} p(y^{*}|x^{*}, w_{n})
\end{equation}
\\
By utilizing this equation, we are able to calculate the predictive entropy from equation \ref{eq:predictive_entropy_I} as
\begin{eqnarray}\label{eq:predictive_entropy_II}
\mathbb{H}_{p}\approx - \sum^{C} \left(\dfrac{1}{N} \sum_{n=1}^{N} p(y^{*}|x^{*}, w_{n})\right) \\
\log\:\left(\dfrac{1}{N} \sum_{n=1}^{N} p(y^{*}|x^{*}, w_{n})\right)\nonumber
.\end{eqnarray}\\
Furthermore, the approximation in equation \ref{eq:marginalization_approximated} can now also be used to determine the aleatory uncertainty from equation \ref{eq:aleatoric_entropy} as 
\begin{equation}
{H}_{a} \approx  -  \dfrac{1}{N} \sum_{n=1}^{N} \left[\sum^{C} p(y^{*}|x^{*}, w_{n}) \log p(y^{*}|x^{*}, w_{n})\right]\nonumber
.\end{equation}


\subsection{Network Design}\label{sec:network_design}
\begin{figure}
\centering
\includegraphics[width=80mm]{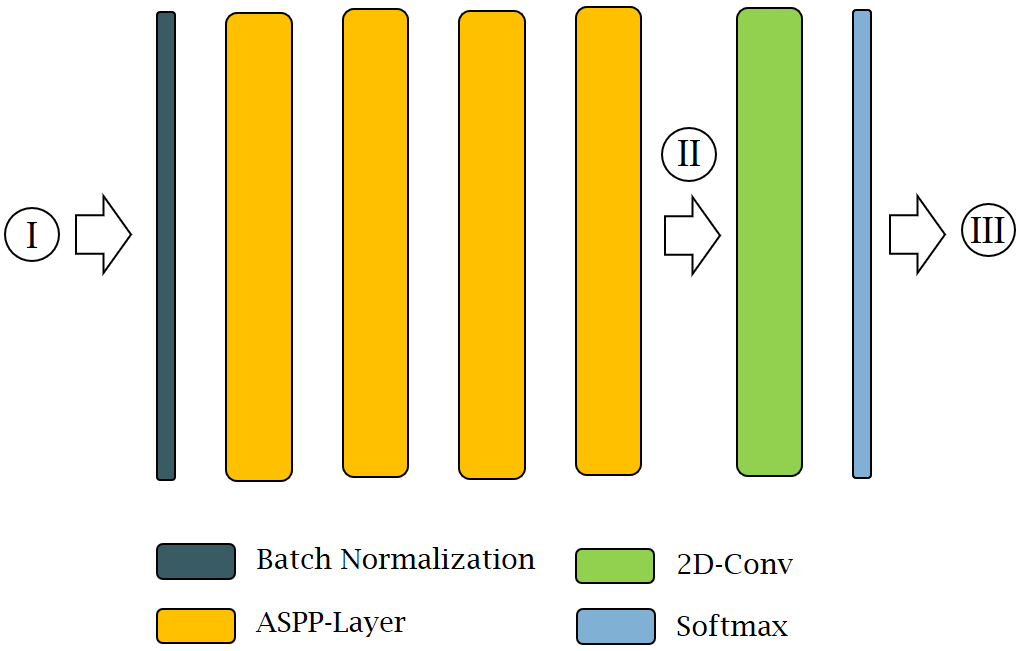}
\caption{\textbf{Network Design:} The $c_{l}$ \textit{x} $c_{w}$ \textit{x} $F_{in}$ input features (I) are batch normalized and then subsequently processed by four ASPP layers \cite{liang_aspp}. In these layers, dilated convolutions are performed with a kernel size of three, stride of one and zero padding so that the grid size stays constant throughout the network. Highly expressive features (II) resulting from the last ASPP layer are then consumed by a 2D Convolution, followed by a softmax function to output \textit{C} class probabilities for each of $c_{l}$ \textit{x} $c_{w}$ cells.
}
\label{fig:network_design}
\end{figure}

The model we deploy is depicted in Figure \ref{fig:network_design}.
After applying batch normalization on the input data, we first utilize Atrous Spatial Pyramid Pooling (ASPP)-Layers introduced by \cite{liang_aspp} to subsequently extract descriptive patterns from the input data: Input features $F_{in}$ containing information about radar reflections (I in Figure \ref{fig:network_design}) are processed to obtain patterns in spatially local correlations of neighboring cells. By deploying a variety of 2D convolutions with different dilation rates in any ASPP layer, feature maps are processed in parallel on different scales to realize a variety of field of views. This allows the network to reliably extract information about objects of different sizes that may be present in numerous neighboring cells. The features that are extracted by parallel convolutions in an ASPP layer are then cell-wise concatenated while maintaining $c_{l}$ \textit{x} $c_{w}$ grid cells. We increase the capacity of the network to recognize patterns in the data by subsequently applying these layers as described in Figure \ref{fig:network_design}.\\
The $c_{l}$ \textit{x} $c_{w}$ high level, complex features as a result of the stacked ASPP layers are then processed to cell-wise output probability scores for each of the \textit{C} considered classes. For this purpose, we use one 2D-Convolutional layer with \textit{C} outputs logits per cell and apply a Softmax-function to obtain a \textit{C}-dimensional categorical distribution.\\
As mentioned in section \ref{sec:epistemic_uncertainty}, we further modify the network in order to capture epistemic uncertainty. Based on the structure presented in Figure \ref{fig:network_design} we implemented three different modifications that will be evaluated and compared in section \ref{sec:experiments}: First, we place parameterized Gaussian distributions as described in Figure \ref{fig:bnn} on all weights of our network. We then introduce a hybrid structure containing deterministic weights in between I and II and Gaussian distributions on the weights in between II and III of Figure \ref{fig:network_design}. In a third configuration, we place a dropout layer on II that remains active during testing to provide a MC dropout implementation of our approach.\\

\section{Experiments}\label{sec:experiments}
In this section, we describe the setup and results of experiments we conducted to verify the ability of our network to reliably perceive the environment while capturing aleatoric and epistemic uncertainties. For this purpose, we first describe how we preprocess input data and GT. To evaluate the effects of the network adaptations that we described in Chapter \ref{sec:method} on the performance of the system, we first train the model from Figure \ref{fig:network_design} in a purely deterministic configuration. We then analyze, both qualitatively and quantitatively, the extent to which quantifying aleatory and epistemic uncertainty allows for more accurate and reliably prediction. We further evaluate various network layouts to capture model uncertainties: A purely probabilistic model in which all weights are replaced by parameterized Gaussian distributions, a hybrid approach of probabilistic and deterministic weights and networks which apply MC dropout as described in current state-of-the-art approaches.

\subsection{Data Preprocessing}\label{sec:data_preprocessing}

To train and evaluate the performance of our approach, we use recordings from a vehicle equipped with radar and lidar sensors. From the radar sensors we obtain point clouds, i.e. detections in two dimensions containing properties that we use as input data to our network.\\
We pre-process the input data by projecting a grid of $c_{l}$ \textit{x} $c_{w}$ cells on the area around the vehicle with the ego vehicle located in the center. In order to generate denser input scans, we concatenate a fixed amount of subsequent point clouds and compensate for motion of the ego vehicle between successive recordings in time. We then assign each detection to its respective cell based on its coordinates $\text{P} = (p_{l}, p_{w})$. Reflections outside the projected grid frame are disregarded. We then project properties from detections in the point cloud to the associated grid cells. Among others, these features include the \textit{amount} as well as the average \textit{Doppler} values, average \textit{RCS} values and the relative \textit{time of recording} of all detections within a grid cell. Additionally, each feature is normalized individually so that the ranges of all input values are within [0, 1].\\
We use annotated lidar point clouds to create cell-wise labels. For lidar reflections resulting from static surroundings, we concatenate all scans within a scene to receive a denser representation of the environment. In order to transfer annotations from three-dimensional lidar point clouds to two-dimensional grid cells, we project the reflections on the x-y plane. A cell is assigned to the class that is most contained in the projected point clouds. Cells that do not contain a single projected lidar reflection are labeled as \textit{unknown}.\\
To focus on the cells that are observable by the radar sensor during training, we apply a cell-wise loss weighting. To this end, we simulate rays between the position of each radar sensor and corresponding lidar reflections. We then derive observability weights by calculating the density of simulated rays per cell relative to the amount of rays that could be present in a cell if they were not blocked by obstacles. As a result, we weight the loss values for cells with high observability with values close to 1, while occluded cells only marginally affect the training due to their loss weighting close to 0.\\
To test the ability of our network to generalize on unseen data, we split our data into a training (\textit{21776} scenes) and a test set (\textit{9294} scenes).

\subsection{Setting}
We trained our models utilizing Adam optimizer with a learning rate of $5$ \textit{x} $10^{-4}$ and a batch size of four. For networks that are equipped with parameterized Gaussian distributions on the weights we applied the loss function defined in equation \ref{eq:kl_loss}, other networks are trained by using the Cross Entropy Loss.\\
We compare the performance of our network on the semantic segmentation task by calculating the Intersection over Union (IoU) for each of the classes \textit{free}, \textit{occupied}, \textit{moving object} and \textit{unknown} and then average over all classes to obtain the mean IoU value (mIoU). Since we only want to consider cells for the evaluation that can actually be observed by the radar sensor, we only calculate the IoU including cells with an observability weight greater than zero. We trained each network for 30 epochs. To generate representative results, we repeated the training five times for each approach and use the mean value over all five training results respectively for evaluation. No data augmentation was applied. 

\subsection{Network Variants}

As a baseline, we first train our model from Figure \ref{fig:network_design} in a pure deterministic implementation, i.e. without utilizing probability distributions on the weights. Per-class mIoU metrics can be obtained from Table \ref{tab:eval_semseg_aspp}. \\
To capture epistemic uncertainty, we then replace each weight in the deterministic network with Gaussian distributions as described in section \ref{sec:network_design}. As a result, we receive a network with distributions $q(w | \theta)$ on the weights that are parameterized by a mean $\mu$ and standard deviation $\sigma$ value, each. By training this network with the loss function defined in equation \ref{eq:kl_loss}, we approximate the posterior probability distribution $p(w | X, Y)$. As shown in Table \ref{tab:eval_semseg_aspp}, this implementation leads to a slightly increased performance of the network measured by mIoU. We attribute this increased performance to the regulatory effect of the probabilistic weights that is caused by the KL divergence loss between the prior $p(w)$ and the approximated posterior $q(w | \theta)$. \\

\begin{table}[t!]
  \scriptsize
  \centering
  \begin{tabular}{ l c c c c c c}
      \hline
      & \multicolumn{4}{c}{Intersection over Union} & \\
      Method & Mean & Free & Occupied & Moving & Unknown & Param. \\
      \hline
      \textbf{Deterministic} & 36.7 & 61.8 & 49.8 & 19.2 & 16.2 & 132.6K\\
      \textbf{Probabilistic} & \textbf{37.2} & \textbf{62.6} & \textbf{51.6} & \textbf{19.8} & 14.7 & 264.4K\\
      \textbf{Hybrid} & 36.7 & 61.6 & 50.1 & 18.6 & \textbf{16.7} & 134.9K\\
      \hline
  \end{tabular}
\caption{\textbf{Semantic Segmentation results (\%)} of the model depicted in Figure \ref{fig:network_design}. The network is implemented both as a consistently deterministic, probabilistic and a hybrid version which is composed of a deterministic feature extraction and probabilistic classification head.} \label{tab:eval_semseg_aspp} 
\end{table}

The implementation of a fully probabilistic network, however, comes at the expense of almost doubling the amount of network parameters without significantly increasing its capacity. The resulting cost increase in terms of computing power and memory is undesired in most domains radar systems are applied. We therefore introduce a hybrid network version that consists predominantly of deterministic network weights and uses parameterized distributions on the weights more deliberately to determine model uncertainty. This model extracts highly descriptive patterns from input data in a deterministic manner (i.e. all weights between I and II in Figure \ref{fig:network_design}). The resulting features in II are then mapped onto the output cells by utilizing Gaussian distributions for all weights between II and III that are able to capture epistemic uncertainties in the convolution operations. We believe that the model uncertainty of the front part can be compensated and reflected to a large extent by restricting the network to use parameterized distributions in the last layers of the network. Thus, parameters $\mu$ and  $\sigma$ that were used to parameterize the Gaussian distributions to capture model uncertainties in the henceforth deterministic layers can be reduced to a single parameter per weight again. As stated in Table \ref{tab:eval_semseg_aspp}, a comparable performance of this hybrid network version is achieved by only marginally increasing the amount of parameters compared to the purely deterministic baseline. This hybrid network can then be used similarly to the fully probabilistic approach to capture epistemic and aleatoric uncertainties as described in section \ref{sec:measuring_uncertainties}. 
\begin{figure}[t]
\centering
\includegraphics[width=70mm]{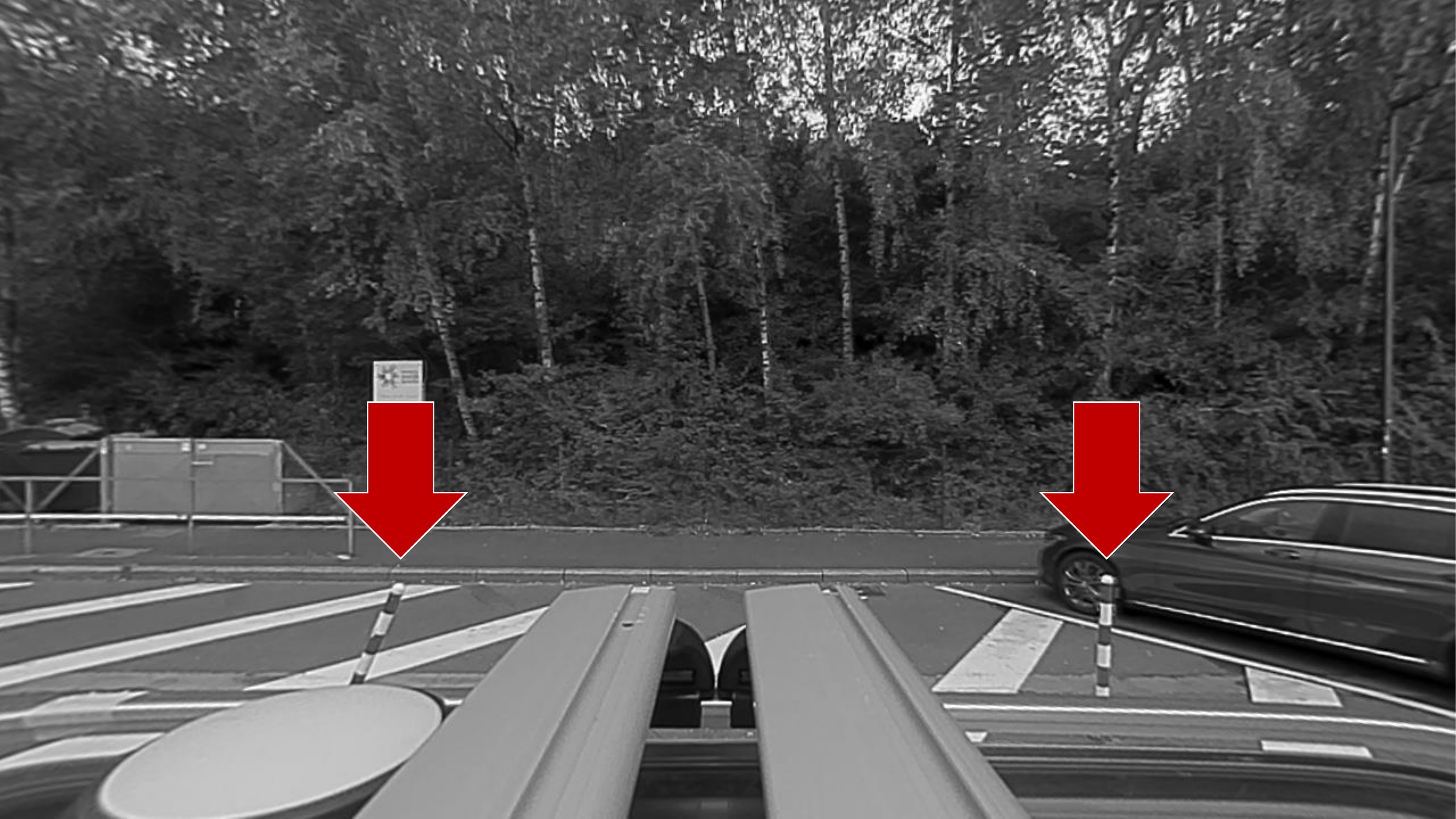}
\caption{Left side view of scene 1 from Figure \ref{fig:eval_graphical}. In this shot, we focus on the two bollards (red arrows) on the left side of the vehicle, which are not present in the GT and are also not recognized by the semantic segmentation network output.}
\label{fig:bollards}
\end{figure}
\begin{figure*}[!b]
\centering
    \includegraphics[trim=60 60 60 60, clip, width=35mm]{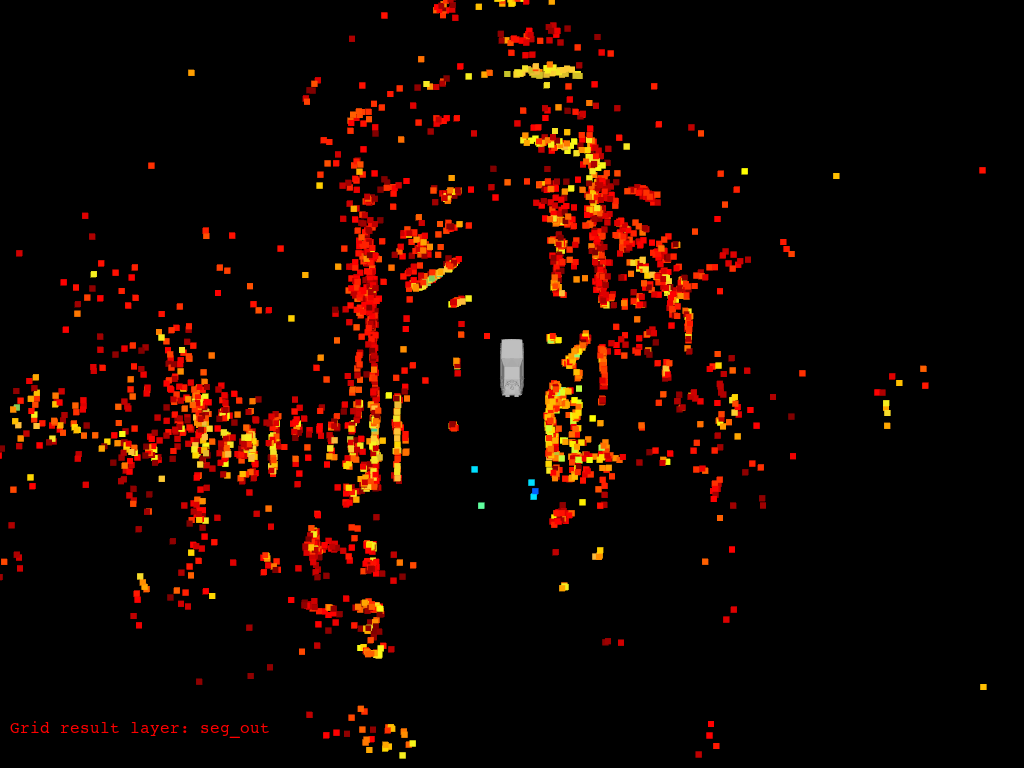}
    \includegraphics[trim=60 60 60 60, clip, width=35mm]{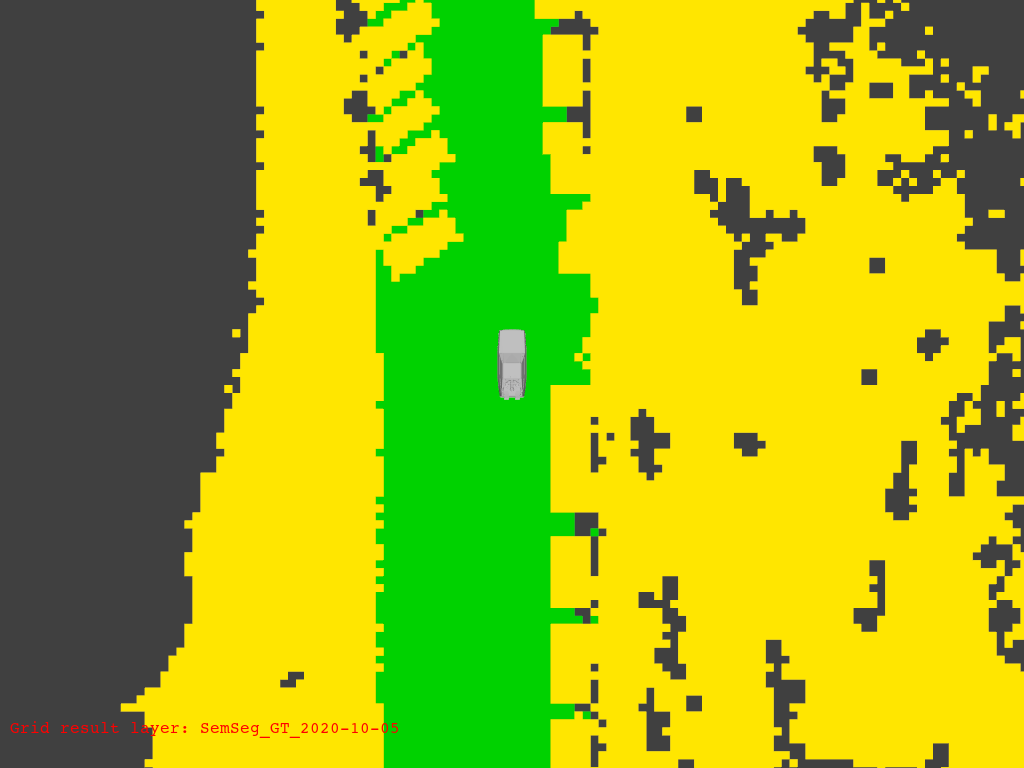}
    \includegraphics[trim=60 60 60 60, clip, width=35mm]{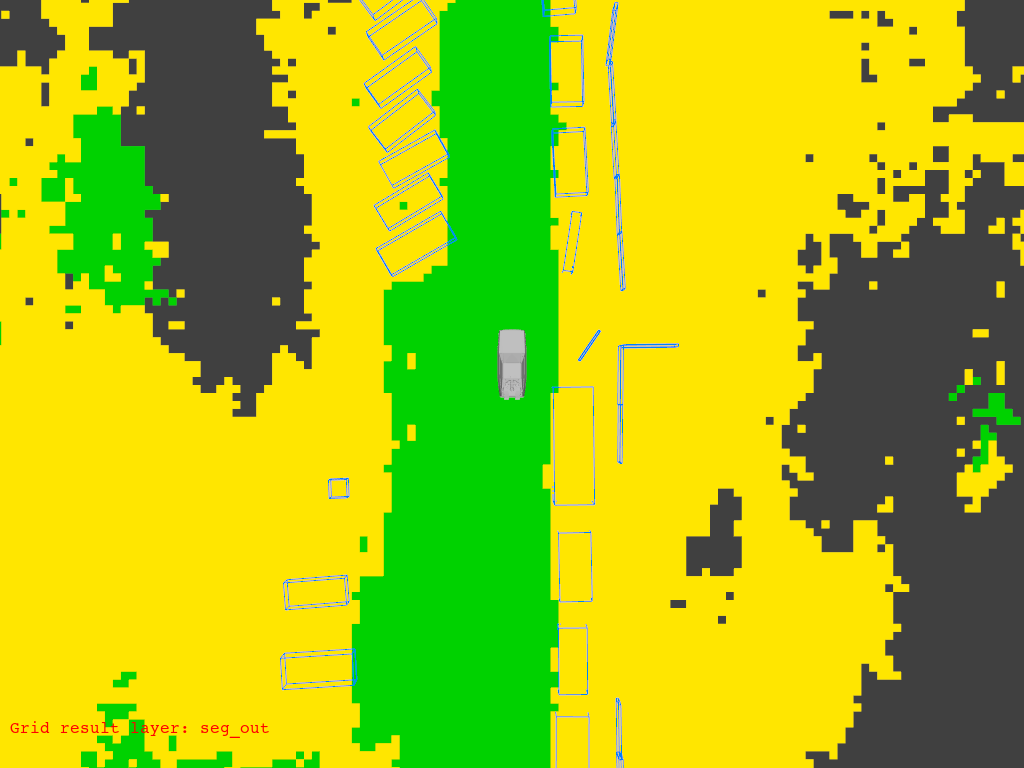}
    \includegraphics[trim=60 60 60 60, clip, width=35mm]{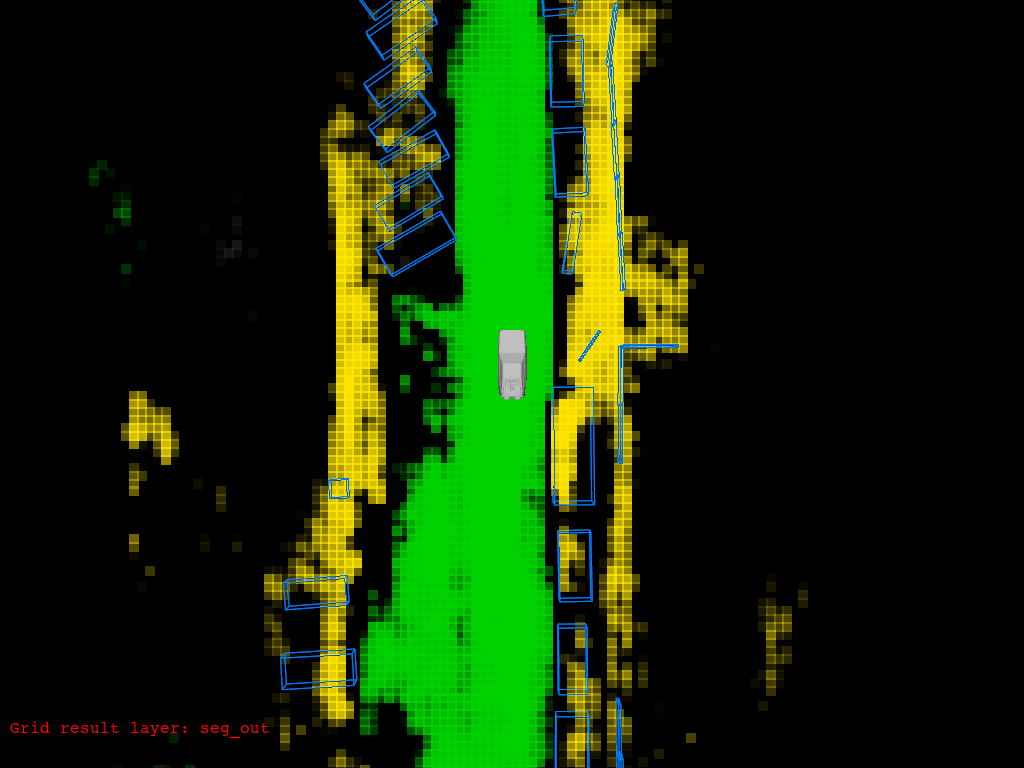}
    \includegraphics[trim=60 60 60 60, clip, width=35mm]{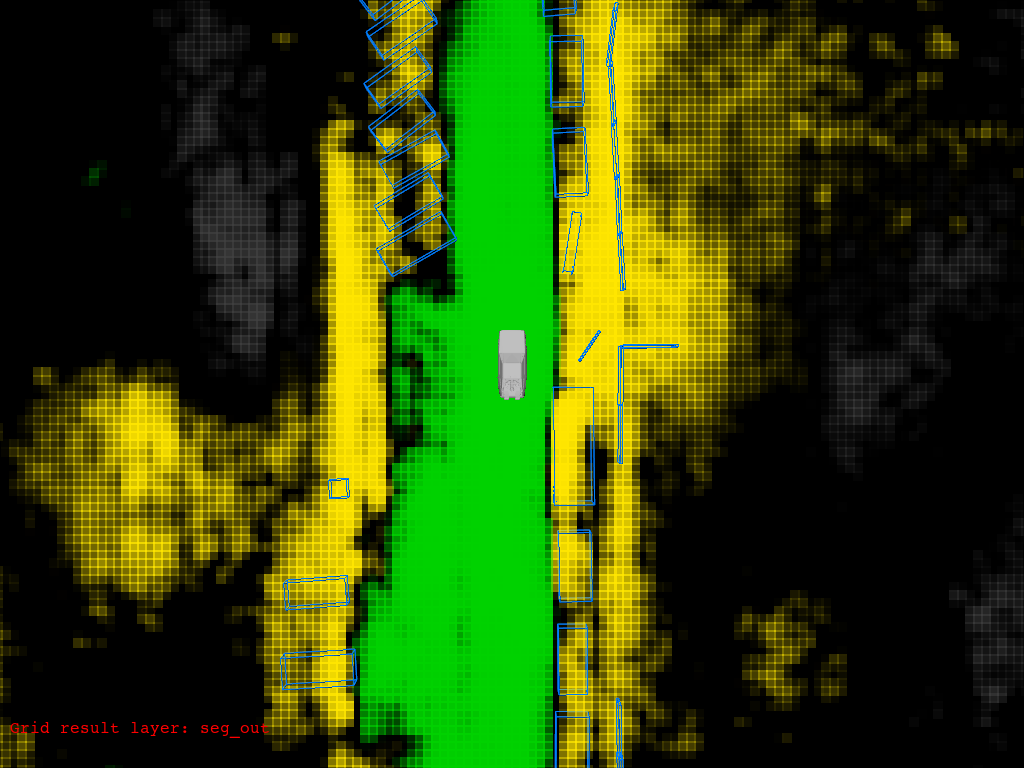}\\
    \vspace{0.1cm}
    \includegraphics[trim=60 60 60 60, clip, width=35mm]{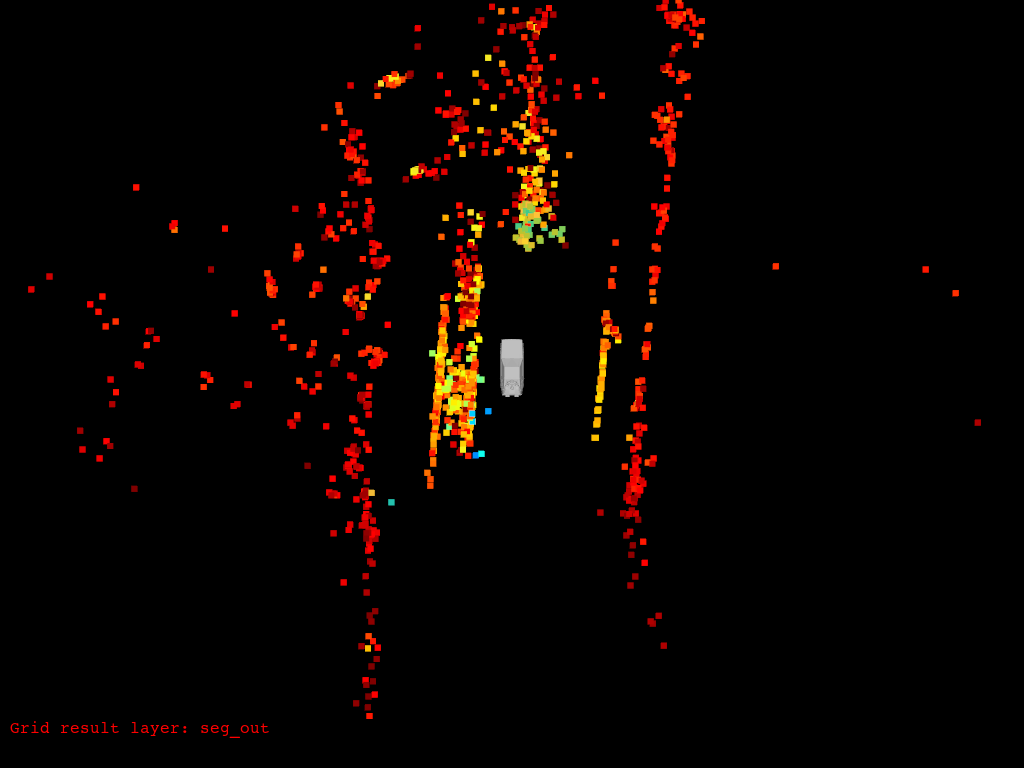}
    \includegraphics[trim=60 60 60 60, clip, width=35mm]{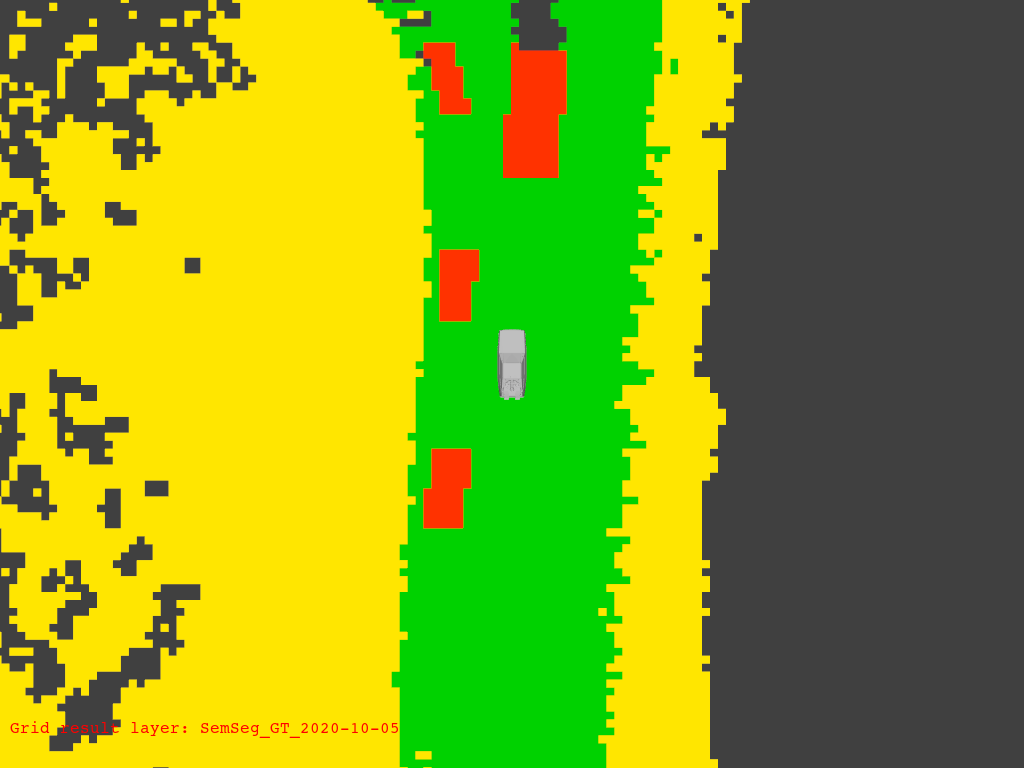}
    \includegraphics[trim=60 60 60 60, clip, width=35mm]{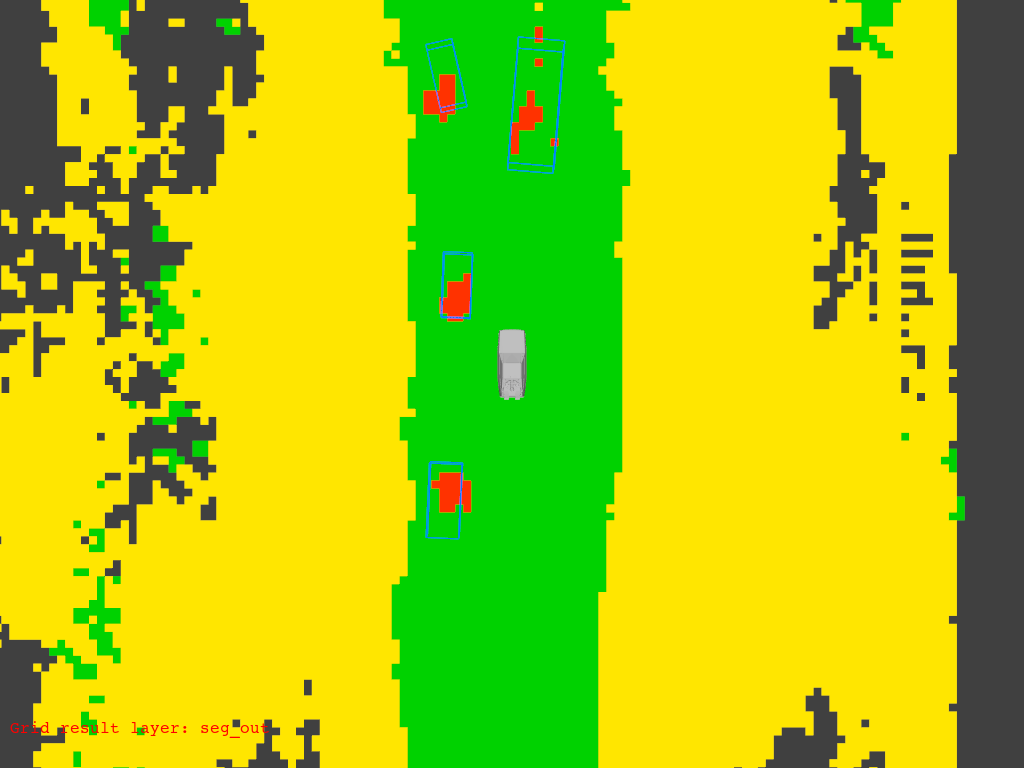}
    \includegraphics[trim=60 60 60 60, clip, width=35mm]{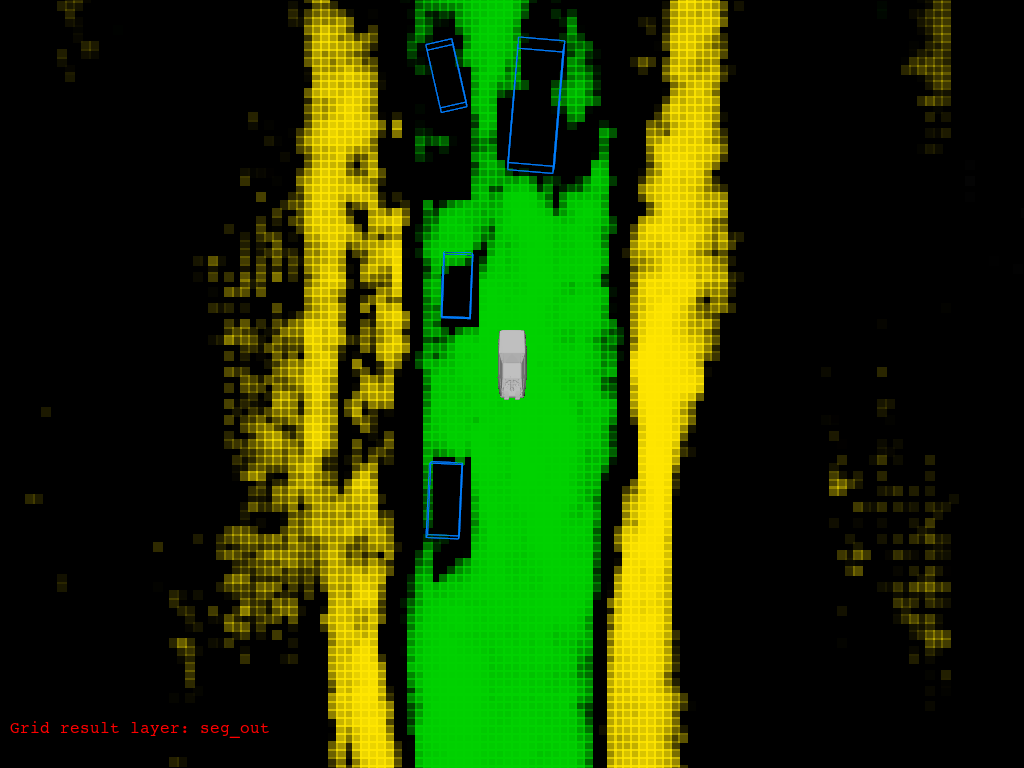}
    \includegraphics[trim=60 60 60 60, clip, width=35mm]{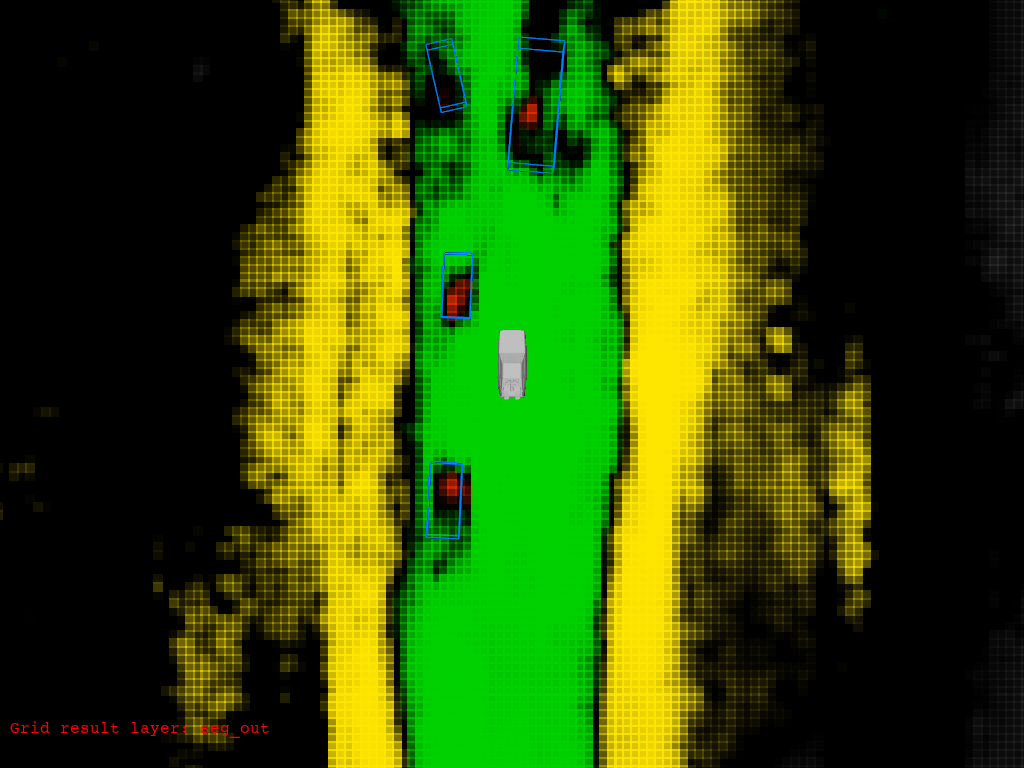}\\
    \vspace{0.1cm}
    \includegraphics[trim=60 60 60 60, clip, width=35mm]{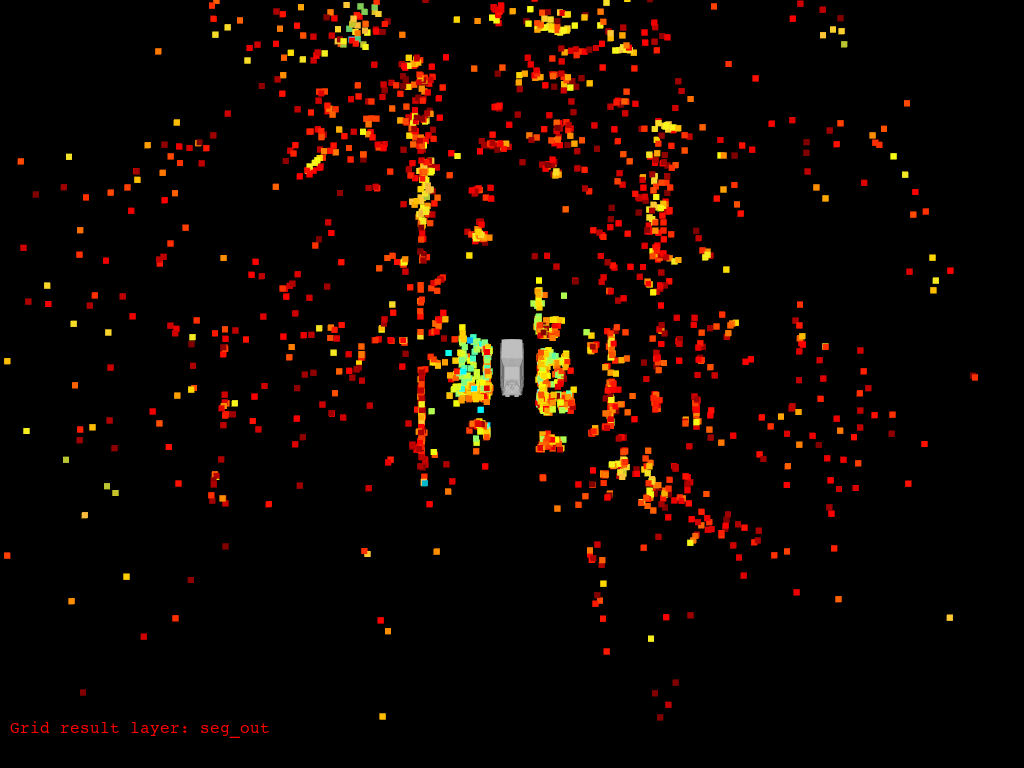}
    \includegraphics[trim=60 60 60 60, clip, width=35mm]{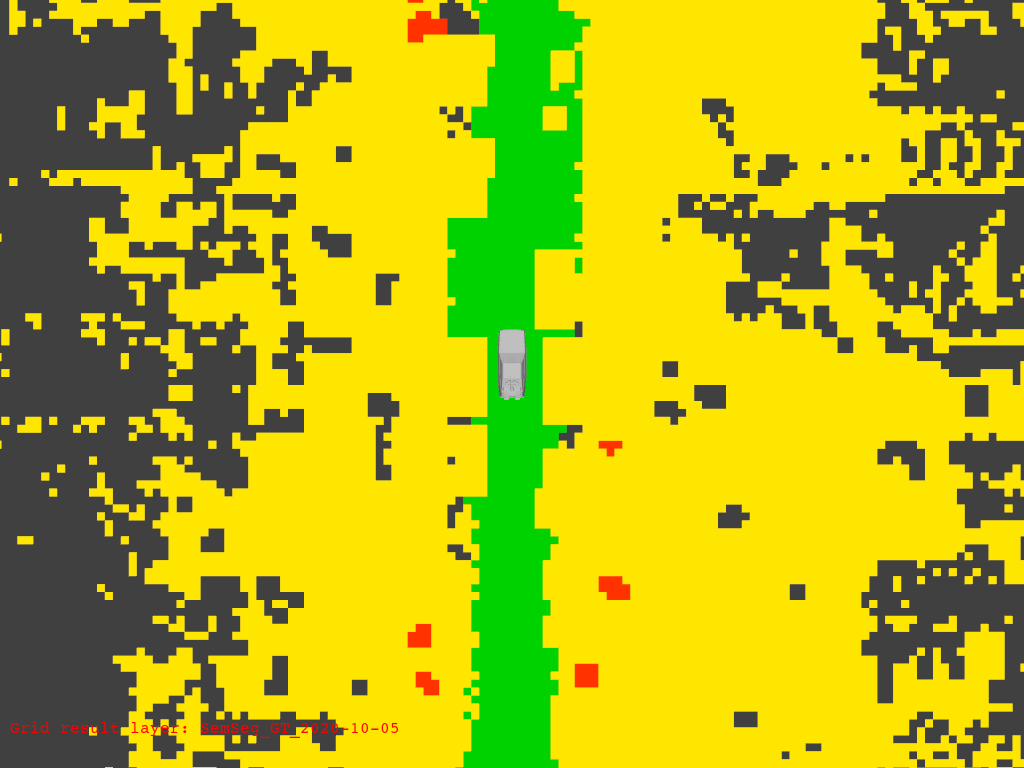}
    \includegraphics[trim=60 60 60 60, clip, width=35mm]{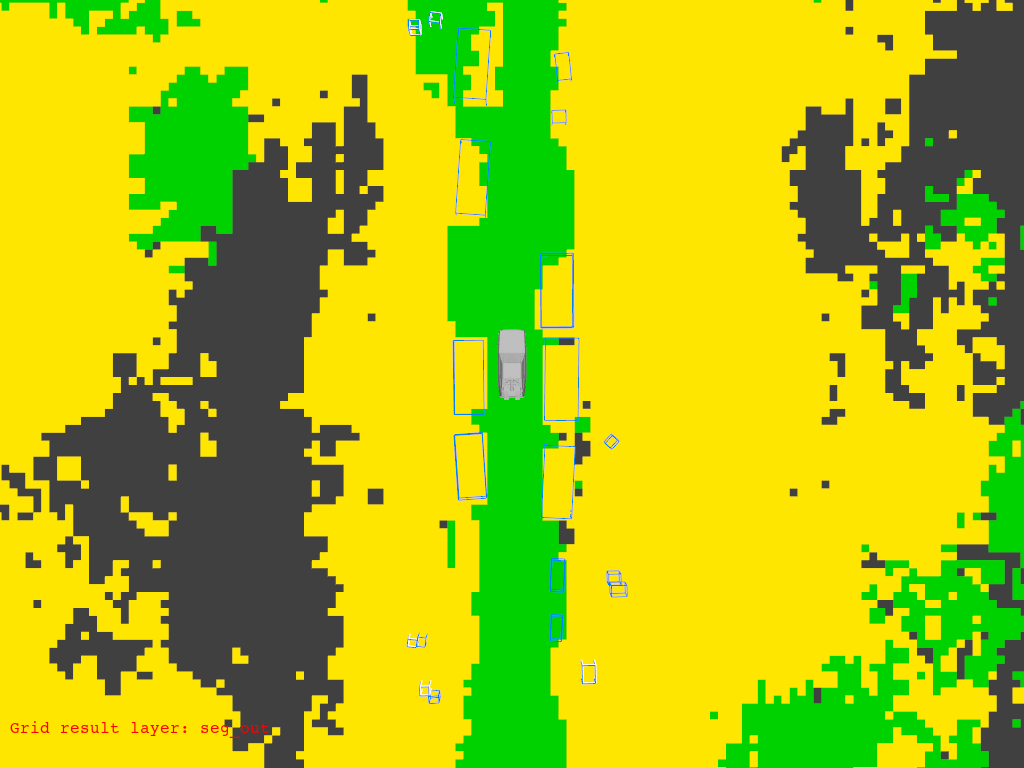}
    \includegraphics[trim=60 60 60 60, clip, width=35mm]{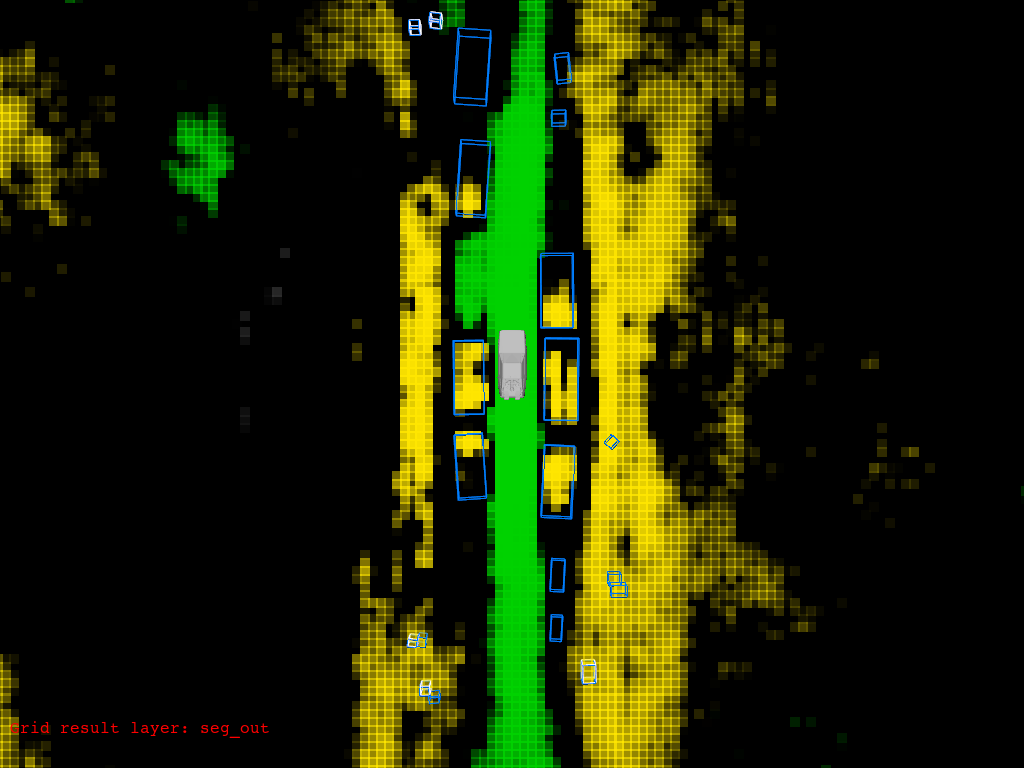}
    \includegraphics[trim=60 60 60 60, clip, width=35mm]{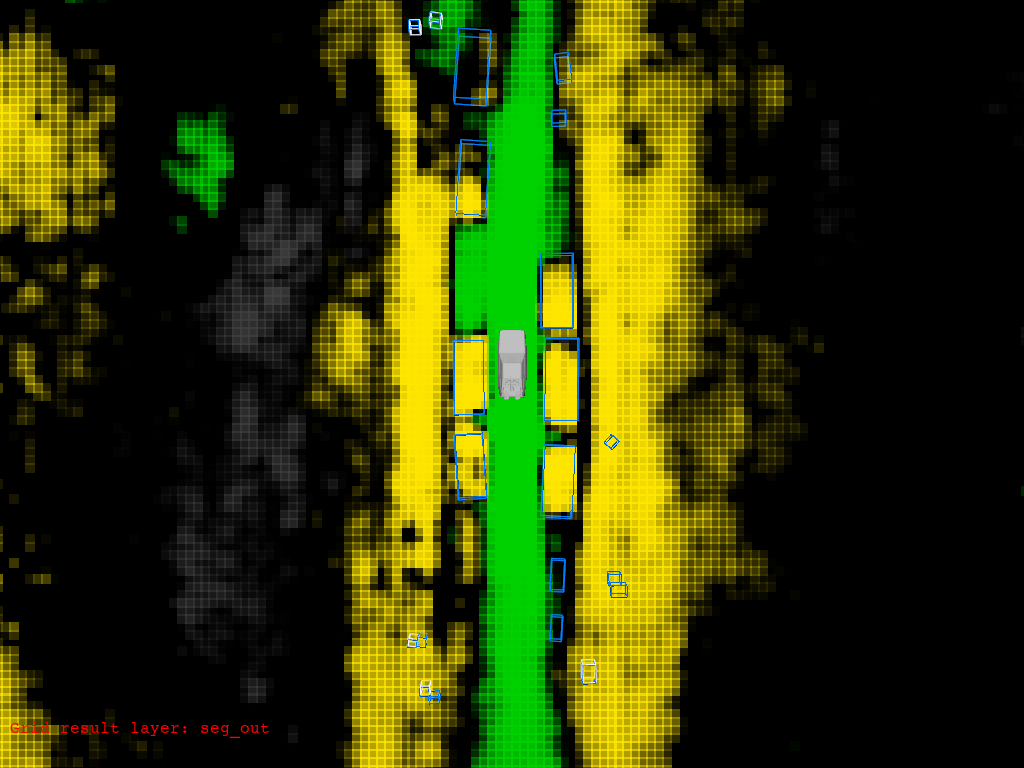}\\
\caption{\textbf{Visualization of semantic segmentation data and results on three scenes}: The left column shows radar point clouds that, transferred to grid cells, are used as input for the network. Red coloring represents the RCS value of each detection. Associated GT data on a cell-basis can be obtained from the second left column. Green cells indicate \textit{free space}, yellow cells \textit{occupied} areas, red cells indicate \textit{moving objects} like vehicles and moving pedestrians and gray cells represent areas that are \textit{unobserved} by lidar. The central column shows the plain output from the network defined in Figure \ref{fig:network_design} in a hybrid deterministic - probabilistic implementation as stated in section \ref{sec:network_design}. Epistemic uncertainty associated with the network predictions is presented in the second right column by obscuring the cells with increasing uncertainty value. Similarly, aleatoric uncertainties for each prediction can be obtained from the rightmost column. In the three right columns, blue bounding boxes indicate objects in the scenes.}
\label{fig:eval_graphical}
\end{figure*}
\subsection{Qualitative Results}
Visual representations of the results from the hybrid deterministic, probabilistic network as described in the previous section can be obtained from Figure \ref{fig:eval_graphical}. The first scene, depicted in the first row, motivates capturing of uncertainties in a network prediction to avoid missing out various objects on areas that are predicted as drivable, i.e. indicating possible false positive \textit{free} cells. In Figure \ref{fig:bollards} two bollards can be observed on the left side of the vehicle that are not present in the GT as shown in Figure \ref{fig:eval_graphical}. Since the GT is derived from concatenated lidar frames, missing out small objects can be caused by a low scan occupancy of lidar reflections on these objects due to their size. Accordingly, the plain semantic segmentation output which is depicted in the third column of the same figure shows that the network will learn to interpret the radar reflections resulting from these bollards as free areas. This can potentially lead to fatal crashes in case an autonomous system operates on this environment perception. This hazard can be resolved by estimating uncertainties in the network prediction as depicted in the last two columns of Figure \ref{fig:eval_graphical}. 
In particular, we observe an increased epistemic uncertainty in the prediction of free cells in the vicinity of the bollards. This shows that although the network predicts the desired class according to the GT, it is able to express its lack of confidence in processing input data that reveals unusual properties for a \textit{free} area. An increased epistemic uncertainty indicates that the network was not trained to predict free cells based on patterns that are similar to those of the detected bollards. Aleatoric uncertainty is less pronounced in this situation, which underscores our assumption that the uncertainty related to these unrecognized bollards arises from a lack of knowledge in the network parameters rather than noise in the data.\\
In each scene from Figure \ref{fig:eval_graphical}, our approach particularly determines an increased epistemic uncertainty for cells that cannot be observed by the radar sensor, for example at the edges of the scene. This effect can be attributed to the cell-wise loss weighting, which, as described in Section \ref{sec:data_preprocessing}, depends on the observability value of a cell. 
Furthermore, all three scenes show that aleatoric and epistemic uncertainties occasionally occur jointly. From this observation we can derive that one uncertainty compensates the other to a certain degree. This property was, to our best knowledge, first mentioned in \cite{Uncertainties_in_CV}.
\begin{figure}
\centering
\includegraphics[width=43mm]{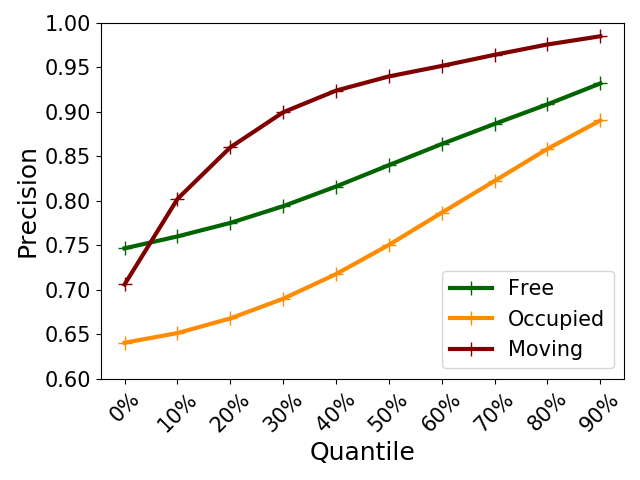}
\includegraphics[width=43mm]{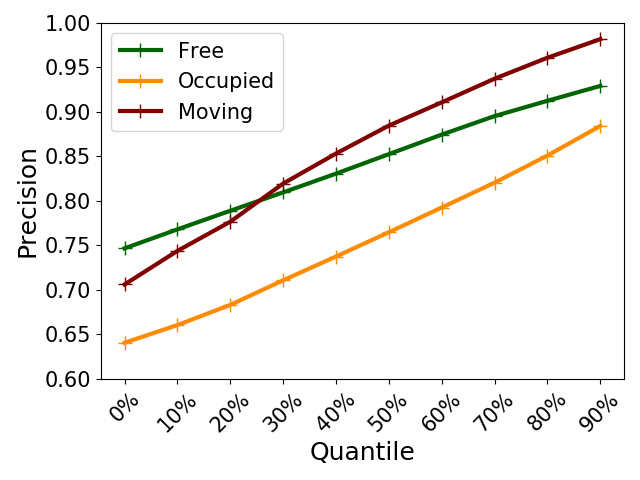}
\caption{Class-wise correlation between network precision and epistemic (left) and aleatoric (right) uncertainties.}
\label{fig:corr}
\end{figure}

\subsection{Quantitative Results}
Besides presenting a visual representation from Figure \ref{fig:eval_graphical} of the network outputs for the hybrid network structure, we evaluate the correlation between certainties of a prediction and the likelihood for a cell to be predicted correctly in the plots of Figure \ref{fig:corr}. For these diagrams we treat each cell prediction of the test set, composed of class probabilities together with epistemic or aleatoric uncertainty, individually. Again, we only consider cells that are visible as stated in section \ref{sec:data_preprocessing}. We then define ten quantiles for both kinds of uncertainty, respectively. By calculating the \textit{precision} of all predictions for each class that are above a certain quantile, we can deduce whether a high cell certainty corresponds to a high probability that a cell will be correctly predicted. \textit{Precision} for each class is defined as the amount of correct classified predictions divided by the sum of correct classified predictions and false classified predictions. By capturing this correlation, Figure \ref{fig:corr} shows strict monotonically increasing graphs for both epistemic and aleatoric uncertainty on a class basis. \\
Based on the results from Figure \ref{fig:corr} we assume that our approach is able to successfully identify predictions on a cell level that are likely to be incorrect based on quantified uncertainties. Since this conclusion holds for both epistemic and aleatory uncertainty, it proves that our approach is reliably able to predict both uncertainty due to an inadequate model and noisy data.

\subsection{Comparison with MC Dropout methods}\label{sec:mc_dropout}
\begin{figure}
\centering
\includegraphics[width=43mm]{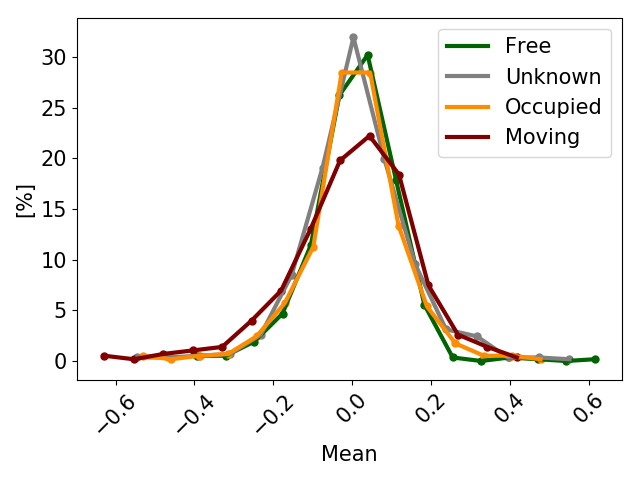}
\includegraphics[width=43mm]{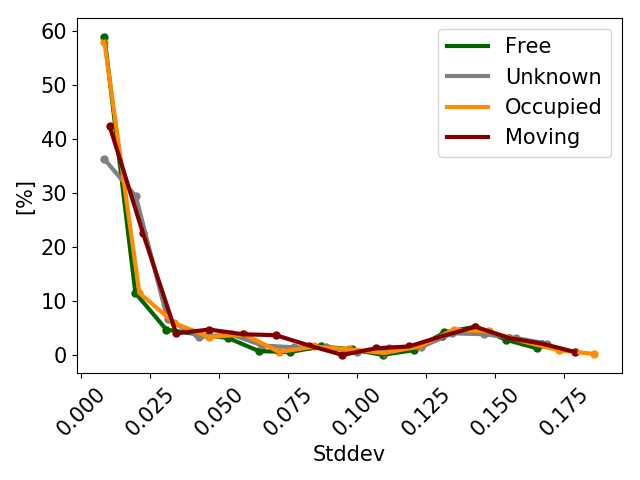}
\caption{Density plots of the mean and standard deviation for parameterized Gaussian distributions from the last layer.}
\label{fig:weight_parameters}
\end{figure}

As stated in section \ref{sec:related_work}, most previous approaches like \textit{Bayesian Segnet} \cite{Bayesian_SegNet} build up on MC Dropout methods \cite{dropout_as_bayesian_approximation}. These approaches capture epistemic uncertainty by implementing dropout layers at certain positions within the neural network that are activated during application of the algorithm. Therefore, subsequent weights are treated as Bernoulli distributions with a fixed probability. Since the probability of switching off a weight is not learned by backpropagation, these MC dropout approaches represent a simplification of BNNs as defined in section \ref{sec:epistemic_uncertainty}. Previous publications like \cite{dropout_as_bayesian_approximation}\cite{Bayesian_SegNet}\cite{Uncertainties_in_CV}, however, show promising results based on MC Dropout methods to estimate epistemic uncertainty in a network. Bayesian Segnet \cite{Bayesian_SegNet} presents a central contribution to this topic. The authors equip \textit{Segnet} \cite{segnet} with dropout layers in various positions within the network and sample through the network during inference to extract model uncertainty for semantic segementation tasks.\\
To relate our results to this approach, we implement \textit{Segnet} as described in \cite{Bayesian_SegNet} in a shallow version to fit the model capacity to our baseline network from Figure \ref{fig:network_design} in terms of trainable parameters. We apply the configuration of \textit{Bayesian Segnet} that performed best as stated in \cite{Bayesian_SegNet} by placing a dropout layer in between encoder and decoder and then apply it for environment perception based on radar data. Results on our test set are depicted in Table \ref{tab:eval_semseg_mc_dropout}. When compared to our network architecture \textit{Gaussian Weights}, we obtain a decrease in mIoU of ${\huge\raisebox{-0.9ex}{\~{}}}$ 3\% for the \textit{Bayesian Segnet}. \\
This difference in network performance, however, can also be related to differences in the model structures between \textit{Segnet} and our implementation. For a more realistic comparison between MC Dropout and the parameterized Gaussian distributions to capture model uncertainties for the environmental perception task, we transfer our network structure described in Figure \ref{fig:network_design} to an MC Dropout implementation. This is achieved by placing a MC Dropout layer in II of Figure \ref{fig:network_design}. While Table \ref{tab:eval_semseg_mc_dropout} shows that this implementation leads to an increased performance compared to \textit{Bayesian Segnet}, learning parameterized Gaussian distributions on the weights of the network still shows a superior performance compared to MC Dropout of 2\% in mIoU.
\begin{table}[t!]
  \scriptsize
  \centering
  \begin{tabular}{ l c c c c c c}
      \hline
      & \multicolumn{4}{c}{Intersection over Union} & \\
      Method & Mean & Free & Occupied & Moving & Unknown & Param. \\
      \hline
      \textbf{Gaussian} & \textbf{37.2} & \textbf{62.6} & \textbf{51.6} & \textbf{19.8} & 14.7 & 264.4K\\
      \textbf{Bayesian Segnet} & 34.8 & 60.2 & 49.2 & 15.4 & 14.4 & 134.9K\\
      \textbf{MC Dropout} & 36.6 & 61.1 & 50.3 & 18.6 & \textbf{16.3} & 132.6K\\
      \hline
  \end{tabular}
  \caption{Comparison of our approach as depicted in Figure \ref{fig:network_design} in a fully probabilistic implementation (Gaussian weights) as well as a MC Dropout layer on II. Furthermore, we utilize Bayesian Segnet \cite{Bayesian_SegNet} in a slightly modified version as specified in section \ref{sec:mc_dropout}. We evaluate on Semantic Segmentation results (\%).} \label{tab:eval_semseg_mc_dropout} 
\end{table}\\
Besides increasing the performance on the semantic segmentation task, applying Gaussian distributions on the weights of the network to capture model uncertainty offers a variety of further advances compared to MC Dropout methods. First, we are able to incorporate external knowledge about class-related epistemic uncertainties into the training by adjusting prior distributions $p(w)$ of equation \ref{eq:kl_loss}. Furthermore, learning the standard deviation parameters $\sigma$ for each weight individually enables the network to learn fine-grained model uncertainties. Figure \ref{fig:weight_parameters} depicts how our network utilizes the opportunity to learn a variety of uncertainties $\sigma$ depending on its individual confidence in the weighting parameter based on the data. Weights that are frequently used and consistently optimized throughout the training are more likely to be parameterized by a low $\sigma$. In order to sufficiently account for epistemic uncertainty we therefore conclude that it is advantageous for the network to independently learn uncertainty on its weights in contrast to fixed Bernoulli distributions as applied in MC Dropout methods. \\
Furthermore, the approach of learned Gaussian distributions on the weights provides increased comprehensibility regarding relevance and reliability of individual network parameters. Weights that are parameterized by a high standard deviation $\sigma$ are thus more likely to be interpreted as less reliable. Approaches such as active learning \cite{Deep_Bayesian_Active_learning} or pruning \cite{Weight_Uncertainty_in_Neural_Networks} can leverage these insights to increase network performance and make networks more efficient.
\section{Summary}
In this work, we implemented and evaluated a neural network architecture to perform environment perception based on radar data as a semantic segmentation task. We further defined weights in the network by parameterized Gaussian distributions that are able to capture model uncertainty to increase reliability and accuracy of network predictions for environment perception. We furthermore differentiate between model uncertainties (epistemic) and uncertainties resulting from noise in the data (aleatoric).\\
Parameterizing Gaussian distributions on the weights of a network, however, doubles the amount of parameters that are utilized by the network. We therefore present a hybrid deterministic, probabilistic network structure that drastically reduces the amount of parameters while remaining its capability to capture model uncertainties.\\
\bibliography{main}{}
\bibliographystyle{unsrt}

\end{document}